\pdfoutput=1
\documentclass[11pt,a4paper]{article}

\usepackage[review]{acl2023}
\usepackage[separate-uncertainty=true]{siunitx}
\usepackage{bm}
\usepackage{times}
\usepackage{amsmath}
\usepackage{amssymb}
\usepackage{latexsym}
\usepackage{booktabs}
\usepackage{multirow}
\usepackage{subcaption}

\usepackage{graphicx}% so it makes black blobs

\usepackage{microtype}
\DeclareUnicodeCharacter{2212}{-}
\author{Arjun Vaithilingam Sudhakar$^{1,2}$, Prasanna Parthasarathi, Janarthanan Rajendran$^{1,3}$, Sarath Chandar$^{1,2,4}$ \\
  $^{1}$ Mila - Quebec AI Institute \\
 $^{2}$ Ecole Polytechnique de Montreal \\
  $^{3}$ University of Montreal \\
  $^{4}$ Canada CIFAR AI Chair\\
  {\tt arjun.vaithilingam-sudhakar@mila.quebec } }
\title{%Tran
Language Model-In-The-Loop: Data Optimal Approach to Learn-To-Recommend Actions in Text Games}
\begin{document}
\maketitle
\begin{abstract}
%Sarath:
% starting is the problem - directly we are jumping
%Jana/Prasanna:

% exisitng paper says, the lesser the finetuning the performance drops
% Analysis of the language model in the loop - study
%Currently, large language models are used to propose action candidates in text-based games. Using a language model to propose the candidate actions does the action pruning. Furthermore, human-annotated data is used for a few shot adaptations. In the existing setups, once the language model is finetuned, it is frozen and used only as a knowledge base to generate candidate actions conforming to the game dynamics learned from the annotated data. Such systems rely heavily on human-annotated data, which are expensive to collect, thereby limiting the transfer to other games or possible real-world scenarios. We hypothesize that the in-game trajectories generated through interactions hold relevant game information that can be leveraged for improved training. To that end, we explore (1) separate buffer and (2) reward \& location-based transition selection as potential solutions. Our proposed approach outperforms $9$ out of $10$ games with an increase of $3.9\%$ of the norm score.  We empirically observe that the proposed methods alleviate the need for human-annotated data. Even with $10\%$ of human data, we surpassed by $1.7\%$ norm score compared to $100\%$ of human data baseline across $10$ games.

Large Language Models (LLMs) have demonstrated superior performance in language understanding benchmarks. CALM, a popular approach, leverages linguistic priors of LLMs--- GPT-2---for action candidate recommendations to improve the performance in text games in Jericho without environment-provided actions.  
However, CALM
adapts GPT-2 with annotated human gameplays and keeps the LLM fixed during the learning of the text based games. In this work, we explore and evaluate updating LLM used for candidate recommendation during the learning of the text based game as well to mitigate the reliance on the human annotated gameplays, which are costly to acquire. We observe that by updating the LLM during learning using carefully selected in-game transitions, we can reduce the dependency on using human annotated game plays for fine-tuning the LLMs. We conducted further analysis to study the transferability of the updated LLMs and
observed that transferring in-game trained models to other games did not result in a consistent transfer. 
\end{abstract}

\section{Introduction}

%Language understanding has a variety of applications in natural language processing(NLP) and is evaluated with challenging benchmarks for text classification \citep{text_classification}, question answering \citep{qa}, \citep{squad}, sentiment classification  \citep{classification}, etc.; In contrast, many conventional classification tasks often evaluate single-step reasoning. Learning on text-based games enables reasoning over multiple steps \citep{experience_grounds_language}. 
%Touch stone of natural language processing(NLP) lies the efficiency in language understanding over the variety of applications. However, such understanding tasks often have variance in the challenges they have to offer. Also, the tasks are generally constructed over publicly available corpora, which also inadvertently used for training large LMs. That affects the evaluation of large models on language understanding tasks that are popular in the NLP literature, there also is a concern of models exploiting the data heuristics that end up casting shadow on the language understanding abilities of the models.

Large Language models \cite{bert, gpt2, instructgpt} (LLMs) trained on large corpora of unstructured text corpora are the state-of-the-art models in several Natural Language Understanding (NLU) benchmarks. \citet{climbing_right_hill} argue in their position paper that the models trained largely from static benchmarks rely to the \textit{form} rather than understanding the meaning. While it is imperative to understand the learning dynamics of LLMs \cite{rogers2020primer,webson2021prompt}, introducing novel language understanding challenges pushes the frontiers for LLMs' applications.
%the benchmarks during the inference does not commonly require adapting to a domain.  However, works like\citep{climbing_right_hill} claim that such non-interactive models trained static datasets. Also, there has not been considerable progress on
There has been a recent interest in interactive training of large language models in situated learning environments. \citet{experience_grounds_language,mcclelland2020placing} point out the necessity for LMs to have enhanced language understanding and meaning through interacting with the physical world. Also, \citet{word_meaning} argues that LMs fall short in their communicative usage, requiring reasoning over intents despite their success in static datasets. 
%(LM) that could be attributed predominantly to the dearth of use cases
%has   pertains to a system that can recognize a user's intent toward solving a task with the entities declared and described by the user
%\citep{climbing_right_hill, experience_grounds_language}.

%While language understanding is expected through learning relationships among entities, and reason over them towards predicting a hypothesis \cite{word_meaning}, \citet{climbing_right_hill} Towards that, modern NLU systems are trained with an elaborate procedure language model(LM) pretraining that requires a system to predict a token based on context \citep{implicit-representation}. 
%have proven to be a helpful setting for agents that can understand and communicate in natural language. 
\begin{figure}[t] 
\includegraphics[scale=0.55]{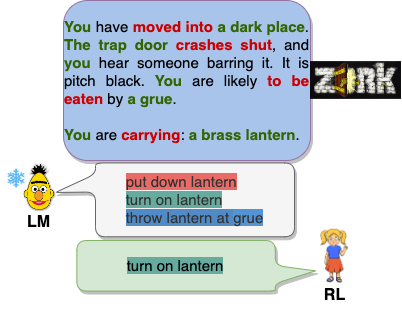}
\caption{Sample gameplay from zork1 game in Jericho using LM for action recommendation: LM recommends action candidates based on the observation from \texttt{env}. The RL agent selects an action from the candidates.
\label{Figure:text_game_example}}
\end{figure}

Training decision making agents over textual information for playing text-based games \citep{interactive_fiction_games, text-world} has been a recent usecase for LLM. While decision making has been the front of text-game playing, such games introduce novel challenges for language understanding, and domain adaptation for LLMs. \citet{ Keep-CALM-Explore-Language-Models-for-Action-Generation-in-Text-based-Games} used GPT-2 \cite{gpt2} to generate candidate actions for the decision making DRRN module \cite{drrn} in Jericho benchmark of text based games. Such a set up allows for qualitatively understanding the LLMs' abilities to \emph{understand}, \emph{reason}, and \emph{adapt} to novel situations. In a typical text-based game, as in \autoref{Figure:text_game_example}, an agent receives a textual observation about its environment that it has to understand and reason over the possible actions to pick one and proceed. While learning from scratch is time-consuming, \citet{Keep-CALM-Explore-Language-Models-for-Action-Generation-in-Text-based-Games} make use of linguistic priors in LLMs to prune the combinatorially large action space. The authors adapt GPT-2 for the task with a corpus of human game play on similar games--- ClubFloyd. After the adaptation phase, the model remains frozen throughout the learning that happens within the game. 
%For successful gameplay, the agent has to handle combinatorial textual action space and operate in a partially observable environment with sparse reward. Hence, the agent should have a solid understanding of the language to progress through these games \citep{}.

% rewrite this paragraph
% Add more citations, more convincing. It's more like an opinion piece. it add's more credibility. 
% use quote when it is critical 
% 
%\citep{interactive_fiction_games}.

Further, \citet{Keep-CALM-Explore-Language-Models-for-Action-Generation-in-Text-based-Games} also note that the performance on the text-based games in Jericho benchmark was sensitive to the size of the annotated human gameplay corpus; such reliance adds to the cost. On the one hand in-game transitions remain unutilized for training the LLM, and on the other there is a need to mitigate the reliance on human annotated transitions to scale applications of LLMs. Although one can make use of the transitions to train the model, the solution requires a comprehensive analysis on what such a LM-in-the-Loop training entails. 
%It is imperative to investigate techniques that reduces the reliance on human annotated data.
Toward that, we explore LM-in-the-Loop by building over the setup in \citet{Keep-CALM-Explore-Language-Models-for-Action-Generation-in-Text-based-Games} by training GPT-2 using in-game generated transitions. Further, we analyze such a set up along the metrics of: (1) Improvement in performance, (2) Acceleration in convergence, (3) Reliance on human annotated transitions, (4) Replacing GPT-2 as a policy network, (5) comparing reward, state based transitions selection for LM training, and (6) Generalization of LM-in-the-Loop trained LM to other games. 
%This allows the LM to also use the data generated by the agent during the game to refine its candidate action generation. Further, such a set up drastically reduces human annotated data to $10$\% to match or perform better than \citet{Keep-CALM-Explore-Language-Models-for-Action-Generation-in-Text-based-Games}. 
The main findings of the approach are summarized as follows:

\begin{itemize}
    \item LM-in-the-Loop reduces emphasis on human annotated transitions and enables accelerated convergence.
    %\item Reducing human annotated data for playing text-based games through \emph{Language Model-in-the-loop}.
    \item State feature based transitions selection provided greater gains than other alternates.
    
    \item LM-in-the-Loop does not always transfer to other games.

    \item Although LM-in-the-Loop improved candidate suggestion, GPT-2 as policy network did faired poorly across games.
    %\item Training LMs (without DRRN) themselves as policy networks within the in-game learning. 
%    \item In addition to that, we created a buffer that holds reward yielded \& agent locations change trajectories. We found that, even with less good trajectory, the agent can still learn better in contrast to \citep{Multi-Stage-Episodic-Control-for-Strategic-Exploration-in-Text-Games} where only high reward trajectories are used.
%    \item Advantage values from the reinforcement learning network as a sample weighted-cross entropy loss for the LM to see if it can help the learning process. Surprisingly, it only helped a little; vanilla cross-entropy was performing equally well. Intuitively, it should have helped, as it gives high weightage to the good action for the state and less weightage to the wrong action.
%    \item To test the LM generalization, we took the LM and continually finetuned it based on the in-game data generated. Then it is used as an initialization for other games to test the game knowledge behavior. We found that if the games are orthogonal, they will take longer or not converge. On the bright side, if there is an overlap between the games, it accelerates the performance. 
\end{itemize}

\section{Related Work}
\paragraph{Text Games:}
Jericho \citep{interactive_fiction_games} is a popular learning environment that supports $32$ human-written interactive fiction games. These games are designed to be difficult for human players, serving as a more realistic training ground to evaluate language understanding agents. Compared with frameworks like TextWorld  \citep{text-world}, these games have significantly more linguistic variety and larger action space. Jericho environment provides a smaller list of candidate actions that can be used to train reinforcement learning (RL) agents. Approaches like DRRN \citep{drrn}, TDQN \citep{interactive_fiction_games}, and KGA2C \citep{kga2c} have used \emph{handicap} to operate on small action space and learn only through in-game rewards. Towards using large LMs, environment provided actions are replaced with LM generated actions like with GPT-2 \citep{Keep-CALM-Explore-Language-Models-for-Action-Generation-in-Text-based-Games}, or BERT \citep{bert_textgame}.

\paragraph{Transformers in RL:} Transformer architectures are now being increasingly used in reinforcement learning (RL); \citet{decision_transformer, rl_one_big_problem} use smaller transformer architectures on Atari games that earlier used convolutional networks as policy networks in offline setting. Further adaptations to make the architectures lightweight to enable online training was proposed in \citet{DRL_with_transformers,stable_transformer,instructgpt, offline-wikipedia, LM_for_offline_rl, grounding_language}.
%[Explain the papers]%add gates and make it lightweight for online training. \citet{decision_transformer} and \citet{rl_one_big_problem} use a smaller transformer and for offline learning.  
 \citet{Keep-CALM-Explore-Language-Models-for-Action-Generation-in-Text-based-Games} explore using the
 semantic prior in GPT-2 for candidate action recommendation in text games. Further, \citet{Multi-Stage-Episodic-Control-for-Strategic-Exploration-in-Text-Games,lm_for_interactive_decision_making} train LMs to remember optimal trajectories to swiftly move to novel game regions. %In our proposed architecture, we expect the %They are inefficient, given that transformers and online reinforcement learning are sample-inefficient. 

%Despite the instability issues in training language generation with reinforcement signals, \citet{fine_tuning_language_model_human_preference} finetune pre-trained GPT-2 models with RL for text generation.  However, learning from non-optimal transitions is challenging. In our work, the objective of the language model is to learn from in-game data for proposing action candidates in a data-efficient way. 
\paragraph{Data Efficiency:}
LLMs \citep{bert_pretraining, few_short_learners} are pretrained with tremendous amount of unstructured text data from \textit{the web} using a generic language modeling objective. Adapting the models to a downstream tasks\citep{qa,squad,text_classification,classification}, however, has been shown to greatly affected by the quality of supervision and the size of the dataset. As reliance on annotated data makes their application hard to scale, techniques like data augmentation \cite{feng2021survey}, using distilled models \cite{radford2018improving}, learning from toyish data \cite{wu2022insights} has been explored has alternatives. However, the approach of making LLMs interactive to be trained in a situated learning environment to reduce the need for annotations is only recently getting popular. %There exist many challenging questions in this largely underexplored domain. 

\section{Background}

\subsection{Text Games} 
In text-based games, at each step $t$, a learning agent interacts with the game environment by generating a textual action $a_{t} \in \mathcal{A}_{t}$ that is relevant to the textual observation $o_t$. The agent receives a scalar reward $r_{t} = \mathcal{R}_{t}\left(o_t, a_t\right)$. 
%The reward discounting $\gamma$ can vary between $[0,1]$ to control the importance of rewards in the distant future. 
%Action  $a$ is taken in latent states $s_t \in S $  will modify the game state. $a$ is admissible at state $s$ if $T(s, a) \neq s$.  Hence, the transition function can be defined as $s' =T(s, a)$.
The agent maximizes the expected cumulative rewards $(r_0, r_1, r_2, \ldots r_N)$, until the end of an $N$-step long episode.
%which is usually until either the agent reaches its goal, it dies, or a maximum number of steps in the game is reached.
%at any given time $t$ by taking a series of actions.
\subsection{DRRN and Advantage Function} 
\label{methodology:rl}

A popular deep RL method used in text-based games is the Deep Reinforcement Relevance Network (DRRN) \citep{drrn}. The observation ($o$) and actions ($a$) are first encoded using separate recurrent neural network encoders (such as a GRU \citep{gru}) $f_o$ and $f_a$ respectively. A decoder $g$ then combines the representations to obtain the Q-value using a network parameterized by $\Phi$:
\begin{equation}
    Q^{\Phi}(o,a) = g(f_{o}(o) ,f_{a}(a)). 
\end{equation}

The DRRN learns to estimate the Q-value through iteratively updating $\Phi$ with experience sampled from a prioritized experience replay buffer with the temporal difference (TD) loss:
\begin{equation} \label{td}
 \mathcal{L}_{TD}(\Phi) = \left(r+\gamma \max_{a' \in A} Q^{\Phi}(o',a') - Q^\Phi(o,a)\right)^2,
\end{equation}
where $r$ and $o'$ are the reward and the observation received after taking action $a$ upon observing $o$, and $\gamma$ represents the discount factor. 
%The next action is chosen based on the softmax sampling of the Q-values.

% The Advantage function calculates how much a173
% particular action is a good or bad decision, given174
% a particular state. It is determined by subtracting175
% the Q-value from the average actions it would have176
% taken in that situation.

\paragraph{Advantage function:} An estimate how good an action, $a$, is when chosen in a state, $o$, is obtained by subtracting the value of the state ($V(o)$)---a weighted average of the Q-values--- from the $Q(o,a)$ of that particular action in that state.
% from the average actions it would have taken in that situation.  
\begin{equation}
    A(o,a) = Q^{\Phi}(o,a) - V^{\psi}(o)
    % \sum_{a'\in\mathcal{A}-a} p(a'\mid o)Q^{\Phi}(o,a'),
    \label{eqn:advantage-function}
\end{equation}

% we approximate $\sum_{a'\in\mathcal{A}} p(a\mid o)Q^{\Phi}(o,a')$ with a value network, $\psi$, that estimates  $\mathrm{V}^{\psi}(o)$. 
Q-Value estimates the expected reward after a specific action was played, whereas $\mathrm{V}^{\psi}(o)$ is the parameterized estimate of the expected reward from being in $o$ before an action was selected.

\subsection{LLM for Action Recommendation} \label{method:lm}

Consider a dataset $\mathcal{D}$ of $N$ transitions of human gameplay across different games organized in context-action pairs as $((\textcolor{blue}{o_{j−1}}, \textcolor{green}{a_{j−1}}, \textcolor{red}{o_{j}} ), a_{j})$. For example: a sample could be like, ``\texttt{[CLS]\textcolor{blue}{$\ldots$  to the north is a restaurant where the mayor ate often. to the east is the mayor's home.} [SEP] \textcolor{green}{northeast}[SEP] \textcolor{red}{$\ldots$ you are carrying nothing.   you are still on the streets. $\ldots$} [SEP] northeast''}. \texttt{[SEP]} and \texttt{[CLS]} are special tokens specific to LM-training. \citet{Keep-CALM-Explore-Language-Models-for-Action-Generation-in-Text-based-Games} uses ClubFloyd to adapt a pretrained GPT-2 model with causal language modeling task. %The  the correct prediction would be ``east``. 
%Training with ClubFloyd \citep{Keep-CALM-Explore-Language-Models-for-Action-Generation-in-Text-based-Games} dataset (Section \ref{dataset}) here aims to inject prior knowledge about the text-based game into the GPT-2 \citep{gpt2} pretrained model and, in return, prune the action space. 
The motivation is to enable the linguistic prior of GPT-2 to adapt to the games and provide better action recommendations to the DRRN.
%recommend to the reinforcement learning agent as it is trained on a massive corpus of data. We train parametrized language model by minimizing the cross entropy loss. Hence, the language model proposes reasonable candidates for text-based games. 
%\citet{Keep-CALM-Explore-Language-Models-for-Action-Generation-in-Text-based-Games} train the LLM with ClubFloyd dataset to adapt and empirically observe that the technique is reliant on the annotated trajectories more than the improvement drastically decreases when only $10\%$ of ClubFloyd dataset is used to adapt GPT-2 model.
%\subsubsection{CALM - 100\% :} \label{baseline:calm100}
%In CALM-100\% variant \cite{Keep-CALM-Explore-Language-Models-for-Action-Generation-in-Text-based-Games}, the whole human-played ClubFloyd dataset is used as a few-shot adaptation on the pre-trained GPT-2 language model.
%\subsubsection{CALM - 10\% :} \label{baseline:calm10}
%In CALM-10\% variant \cite{Keep-CALM-Explore-Language-Models-for-Action-Generation-in-Text-based-Games}, only $10\%$ of the human-played ClubFloyd dataset is used to the pre-trained GPT-2 language model as a few-shot adaptation.

\section{Methodology}

% 2 Baselines 100 and 10 percent

% $p_{\theta}$ to generate actions conditioned on contexts c, and The cross-entropy loss \ref{formula:cross_entropy} is used to learn the model's parameters.

%Unsupervised learning is used as a pretraining objective for finetuning models in a supervised fashion. This technique has proven helpful in avoiding two significant bottlenecks in improving NLP benchmarks: limited annotated datasets and generalization across a wide range of tasks. \citep{gpt2} proposed learning a generative language model with unlabeled data and then finetuning the model with examples of specific downstream effects. 

% Add example

%The ClubFloyd dataset introduced in \citet{Keep-CALM-Explore-Language-Models-for-Action-Generation-in-Text-based-Games} is a collection of human-played transcripts of 590 games. 

% Continually finetune the language model using a replay buffer
% we have 2 replay buffer and samples are drawn randomly from certain percentage
% we fix the amount of gradient update
% Frequency of LM Finetuning
% percentage of important state transition
% what is considered as state tranisition
% 5 language model for trained on 5 different random seeds.

% Add more content - Figure / Algorithm
% cycle explain
% good buffer and random buffer 
% selecting based on the probability
% hyper-parameters
% loss function
% 
\subsection{LM-in-the-Loop to recommend Actions}
\label{sec:techniques}
\begin{figure*}[t]
    \centering
    \includegraphics[width=\textwidth]{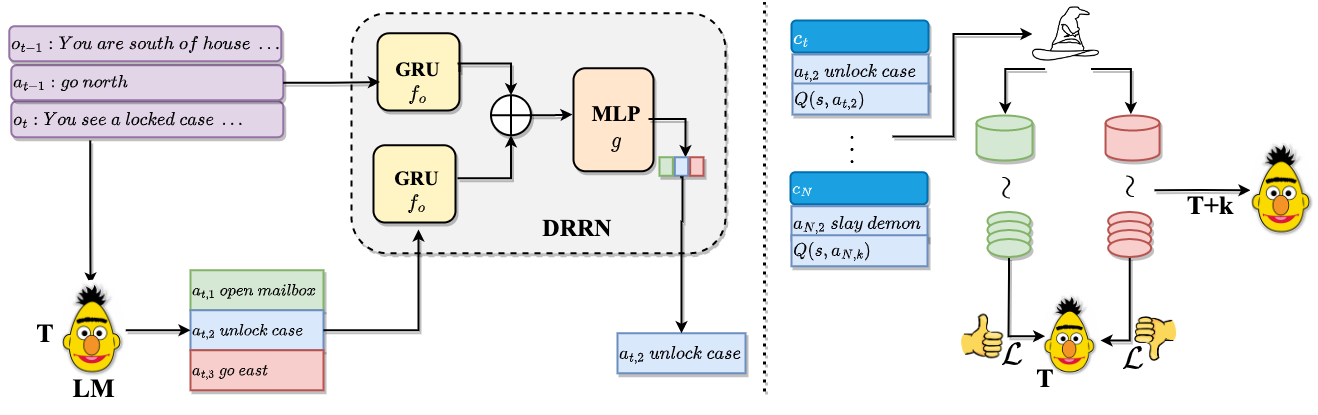}
    \caption{ Training LM-in-the-Loop post-human-annotated dataset adaptation: RL agent (DRRN) picks the action recommended by the language model (at $T$), which is GPT-2. The context pairs are stored in the replay buffers that are categorized by some heuristic. Then the Language model is updated with in-game transitions after $k$ learning steps in the game. Finally, the updated language model ($T+k$) actions are recommended.}
    \label{fig:my_label}
\end{figure*}

The game playing agent follows trajectories that are rewarded according to the rules of the game in the Jericho environment. The environment has two scenarios---with and without \emph{handicap}---which correspond to whether the actions can be generated from within the possible actions suggested by the environment or without any limitations by the environment respectively. The \emph{with handicap} set up  evaluates the agent exclusively on planning with the actions provided, while the \emph{without handicap} requires the agent in addition to understanding the observation also generate acceptable candidates. In \citet{Keep-CALM-Explore-Language-Models-for-Action-Generation-in-Text-based-Games}, the LLM is kept constant throughout the gameplay and that assumption could be only validated if Jericho games share significant similarity with the transitions in ClubFloyd. However, \textsc{Mauve}-Score\footnote{\textsc{Mauve}-Score \cite{mauve} measures  semantic relatedness of an LM generated text with that of human generated text distribution using an LLM representation of the texts.} between the human transitions and the game transitions in Jericho did not overlap significantly (\autoref{tab: mauve-jericho-clubfloyd} in \S\ref{sec:mauve-clubfloyd-jericho}), suggesting that the adapting from in-game transitions is needed. 

Toward that, we explore the feasibility, prospects, and challenges that entail training LM-in-the-loop post finetuning with human gameplays in ClubFloyd adaptation as in \autoref{continual_finetuning}. We use a similar set up for action recommendation as in \citet{Keep-CALM-Explore-Language-Models-for-Action-Generation-in-Text-based-Games}, where a pretrained GPT-2 LM is adapted with Clubfloyd dataset to recommend actions to DRRN agent. In addition to training the DRRN agent with TD-learning (\autoref{td}), we collect the transitions $\left(o_t,a_t,o_{t+1}, r_{t+1}\right)$ throughout the game episode, $e^{TD}$, and populate them in $\mathcal{D}^{+}$ and $\mathcal{D}^{-}$ based on a heuristic that depends on---reward, return, and the game states.

First, with LM parameterized by $\theta$ and generating action candidates, we train DRRN for $n^{RL}$ consecutive episodes. After $n^{RL}$ episodes, we sample $d^{LM}$ sized dataset from $\mathcal{D}^{+}$, and $\mathcal{D}^{-}$ with probabilities $p^{+}$ and $1-p^{+}$ respectively for $2000$ gradient steps at finetuned after every $k$ game steps.  To train LM we use a weighted cross-entropy loss:
%as in \autoref{eqn:cross_entropy}:

\begin{equation}
\mathcal{L}^{LM}(\theta) = -\mathbb{E}_{(a_t,o_t) \sim \left(\mathcal{D}^{+},\mathcal{D}^{-}\right)} \log P_{\theta}(a_t \mid o_t )\cdot h\left(\cdot\right)
\label{eqn:cross_entropy}
\end{equation}

Then, we plug-in back the in-game trained LM to recommend actions for the DRRN agent. The maximum buffer size of $\mathcal{D}^{+}$, $\mathcal{D}^{-}$, $p^{+}$, $d^{LM}$, and $n^{RL}$ are all game-specific hyperparameters. The $h\left(\cdot\right)$ is defined as a function of reward $r_t$, or action-advantage, $A(o_t,a_t)$, or assumed $1$ uniformly $\forall (o,a) \in \mathcal{O}\times\mathcal{A}$. 
We evaluate different approaches based on the sampling of transitions, and the loss function ($\mathcal{L}$), used for training the language model. Approaches for LM-in-the-Loop based on the construction of $\mathcal{D}$, and sampling are:

\paragraph{\emph{Un}categorized Transitions (UT):} In this setting the transitions stored in the buffer are not categorized by any special heuristic function. We simplify this approach by maintaining a single buffer, $\mathcal{D}$ in place of two. This is a weaker baseline than other heuristics to select useful transitions based on their importance.

\paragraph{State Feature Categorized (OC):} In this, the transitions are labeled as useful or not based on whether an action $a_t$ resulted in reward increase or if the agent's location changed. \emph{i.e.,} moved from one room to another. As the location information received is an artifact of the game framework, we consider this as the \emph{Oracle}. Further, we vary $p^{+}$ to maximize the transitions that encourage exploration to eventually result in improved performance in the game. Here, $h\left(\cdot\right)$ is fixed as $1$ uniformly $\forall (o,a) \in \mathcal{O}\times\mathcal{A}$.

\paragraph{Reward Trajectories (RT):} The reward from transitions, $r_t$, is used to categorize positive and negative trajectories. When $r_t > 0$ all transitions up until the earlier non-zero reward are considered positive and added to $\mathcal{D}^{+}. $%, and when $r_t < 0$ to $\mathcal{D}^{-}$. 
% [value-based] Further, we also explore sampling [based on the value] 

Further, we explore utilizing the return, reward, and advantage function of actions to re-weight $\mathcal{L}^{LM}$ using the $h\left(\cdot\right)$ function over {\bf UT} setting as above. We describe them as follows:

\paragraph{Weighted Cross-Entropy:} 
In this, the transition data is kept in a single buffer $\mathcal{D}$ similar to in the $\mathrm{\bm{UT}}$ setting. To finetune the language model using the weighted cross-entropy loss (\autoref{eqn:cross_entropy}), we use the exponential weighted %advantage fu
%g 
advantage function (\autoref{eqn:advantage-function}). We use two variants to the weights, wherein $\mathrm{\bm{UT}}^{EA}$ is non-negative using $h(\cdot)$ function:

\begin{equation}
h(o_t,a_t)  = e^{\beta\cdot A(o_t,a_t)},
\label{eqn:weighted-ea}
\end{equation}
where, $\beta \in \mathbb{R}^{+}$ is a hyperparameter. The other variant, $\mathrm{\bm{UT}}^{LA}$, allows for negative weights with $h(\cdot)$ as follows:
\begin{equation}
h(o_t,a_t)  = 1+ \beta\cdot A(o_t,a_t),
\label{eqn:weighted-la}
\end{equation}
where, $\beta \in \mathbb{R}^{+}$ is a hyperparameter.
\section{Experiments} 
\label{sec:experiments}
We perform comprehensive experiments \footnote{The codebase for all experiments will be released after the anonymity period.} with LM-in-the-loop set up to study the following questions:
\begin{enumerate}
    \item Does including the language model in the training loop improve performance?
    \item Does LM-in-the-Loop mitigate the reliance on human gameplay transitions?
    \item Should the transitions be categorized for improved learning?
    \item Can we make LM itself a policy network without DRRN with LM-in-the-Loop?
    \item Does training LM-in-the-Loop affect generalization to other games?
\end{enumerate}

\subsection{Task Adaptation Dataset} \label{dataset}

ClubFloyd dataset \citep{Keep-CALM-Explore-Language-Models-for-Action-Generation-in-Text-based-Games} is a collection of crawled data from the ClubFloyd website. The dataset comprises of gameplay from experienced players; however, they may not be familiar with the particular games. The data is preprocessed and contains around $217K$ pairs of context an in the form of $((o_{j−1}, a_{j−1}, o_{j} ), a_{j})$.  
%The vocabulary size of the dataset is $39,670$ with an average trajectory length of $360$.
%We split the ClubFloyd dataset into a 90\% training set and a 10\% validation set for training the language model with pre-trained weights. 
%To make the language models learn to generate relevant actions to the text-based game, they are optimized with the ClubFloyd dataset. 
\subsection{Benchmark and the Metric} \label{evaluation}
% have to add why those 10 games selected - game complexity
Jericho \citep{interactive_fiction_games} is a learning environment that supports human-written interactive fiction games as described in \autoref{Figure:text_game_example}. We chose $10$ games based on the diversity in the challenges faced in each game such as large action space, solution length, and reward sparsity as mentioned in \citet{interactive_fiction_games}. We use the average of the last $100$-episodes' score with standard error for individual games \citep{interactive_fiction_games} as our metric for evaluation.% This score tells how much the agent could progress through the game's objective. 

In  addition, we report the average score normalized (avg. norm) against the maximum score possible in each of the games, which estimates the human-machine gap in text-based games. Finally, we also report the relative performance percentage difference between the baseline and the best approach mentioned as $\Delta \%$ in \autoref{continual_finetuning} to capture the improvement as the range of the scores in each game is different.

% can provide a walk-through example here

\subsection{Model Details}
\label{sec:model-training-details}
% Lastly, we present that the reinforcement learning agent can improve performance using less human-annotated data and in-game learning. 

Language model (GPT-2) is first finetuned on %used to train on 
ClubFloyd dataset. Given the context, $(o_{j−1}, a_{j−1}, o_{j})$, the finetuned GPT-2 proposes action candidates for DRRN to choose.
%by conditioning on the context pairs as $((o_{j−1}, a_{j−1}, o_{j}), a_{j})$. 
Following that, every action candidate and context is encoded with a GRU. Then a decoder combines the representations to estimate the Q-value using a multilayer Perceptron (MLP) and updates the DRRN agent parameter $\Phi$.
% language Model
% DRRN
% LM in the loop
% seed value
% buffer size varying
% trust in the experiments
During the training process of the DRRN agents, the context-action pairs are stored in the replay buffers. After $k$ steps, we sample $d^{LM}$ sized dataset from $\mathcal{D}^{+}$, and $\mathcal{D}^{-}$ with probabilities $p^{+}$ and $1-p^{+}$ respectively and update the language model with in-game transitions. Then, the updated language model is used to propose the action candidates.  
%Code implementation is available in GitHub \footnote{https://github.com/chandar-lab/calm/tree/dev}

The buffer size is defined as $100K$ for replay buffers that uses First-In-First-Out (FIFO) strategy to replace samples. To train, $d^{LM}$ samples are sampled uniformly at random from the two buffers $D^+$ and $D^-$. However, the probability of choosing the buffers are defined by $p^+$ and $p^-$ ($1-p^+$) respectively. 
%For example, suppose the $p^+$ is $0.3$. In that case, 30\% of the samples are drawn $D^+$, and 70\% samples are drawn from $D^-$ with a batch size of $1$ and $3$ epochs to finetune the language model as a similar setup explained in the section \ref{lm setup}.  
The number of gradient steps for LM training is fixed at $2000$ across the set ups. 
%So, that the language model always has the same amount of gradient update. 
And, across games we experiment with the hyperparameter $p^+ \in [0,1]$ in $0.1$ increment, and the value for LM finetuning frequency $k \in [2k,5k,10k,20k]$. The results tabled are estimated from $5$ runs.

% \begin{enumerate}
%     \item Use \autoref{fig:my_label} to define the model, training details, and the dataset. - done
%     \item try to distill only the important information about the model from {\bf Language Model Set up} and {\bf Reinforcement Learning Agent Set up} sections.
%     \item Use the same notations as in the Methodology section.
%     \item Flag any missed definition and we can fix it.
%     \item move the detailed model description to Appendix.
% \end{enumerate}

\section{Results}

% \begin{figure*}
% \centering
%     \begin{subfigure}{.32\textwidth}
%   \centering
%   % include first image
%   \includegraphics[width=\textwidth]{Figures/ut.png}  
%   \caption{Uncategorized Trajectories}
%   \label{fig:sub-first}
% \end{subfigure}
% \begin{subfigure}{.32\textwidth}
%   \centering
%   % include second image
%   \includegraphics[width=\textwidth]{Figures/rb1.png}  
%    \caption{Reward Based }%($\mathrm{\bm{UT^{EA}}}$)}
%   \label{fig:sub-second}
% \end{subfigure}
% \begin{subfigure}{.32\textwidth}
%   \centering
%   % include second image
%   \includegraphics[width=\textwidth]{Figures/oc1.png}  
%   \caption{Oracle Categorization}
%   \label{fig:sub-oc}
% \end{subfigure}
% \caption{We compare the acceleration in performance with the different in-game learning techniques. Despite OC achieving better acceleration and performance than the alternatives, we see that other techniques as well provide an accelerated convergence over the baseline. Scores are averaged across 5 seeds}
% \label{fig:acceleration-all}
% \end{figure*}
We follow the questions enumerated in \S\ref{sec:experiments} to analyze the effect of in-game learning of language models for action recommendations.

\begin{table*}[ht] 
\centering
\small
\begin{tabular}{ll|cllcc|c|p{0.5cm}}
\toprule

\textbf{Games} & \textbf{CALM} & \textbf{UT}&  $\bm{\mathrm{UT^{LA}}}$ &  $\bm{\mathrm{UT^{EA}}}$ & $\bm{\mathrm{RT}}$  & \textbf{OC}  & \textbf{$\Delta$$(\%)$}&  \footnotesize{\textbf{Max Score}}\\
 % & \textbf{100\%} & \textbf{100\%-NMS} & \textbf{100\%-Rew. Traj}  & \textbf{100\%-IT}& & \textbf{Score} \\
 % &\ref{baseline:calm100} & \ref{proposed:no_selection} & \ref{proposed:important_transition100}& & \\
\midrule
Zork1 & $30.7_{[4.8]}$  & $32.6_{[4.4]}$ & 
 $30.4_{[8.5]}$ &  $35.6_{[5.7]}$ &  $30.7_{[3.8]}$ & $\bm{38.0}_{[1.7]}$ & $23$\% & $350$ \\
 
Inhumane &	$24.8_{[2.7]}$ & $21.9_{[5.24]}$ &  $28.9_{[11]}$ &  $27.3_{[3.1]}$ & $29.1_{[12.7]}$ &	$\bm{43.4}_{[3.8]}$ & $75\%$  & $90$ \\

Detective	& 	$\bm{290.9}_{[2.7]}$ & $288.5_{[1.5]}$ &$289.3_{[0.2]}$ & $288.3_{[1.3]}$ & $285.1_{[5.6]}$  	& $288.5_{[1.5]}$ & $0$\%  & $360$ \\

Zork3 & $0.3_{[0.09]}$ & $0.3_{[0.14]}$  & $0.4_{[0.1]}$  & $0.6_{[0.1]}$ & $0.6_{[0.1]}$ 	& $\bm{0.7}_{[0.2]}$ & $133\%$  & $7$ \\

Omniquest &		$6.7_{[0.3]}$ & $6.0_{[0.6]}$ & $6.6_{[0.9]}$ & $6.6_{[1]}$ & $6.0_{[0.79]}$  &	$\bm{7.8}_{[1.7]}$ & $16\%$  & $50$ \\

Library	& $11.2_{[1.3]}$ & $9.3_{[1.1]}$  & $9.5_{[1]}$  & $10.3_{[0.2]}$ & $10.3_{[1.8]}$ 	&	$\bm{12.1}_{[0.7]}$ & $8\%$   & $30$ \\

Balances	& $9.3_{[0.2]}$  & $9.6_{[0.1]}$  & $9.6_{[0.2]}$  & $9.5_{[0.2]}$ & $9.7_{[0.2]}$ &	$\bm{9.7}_{[0.1]}$ & $4\%$   & $51$ \\

Ludicorp &  $10.4_{[0.7]}$ & $11.4_{[2.6]}$ & $12.5_{[1.1]}$  & $11.9_{[2.6]}$  & $11.3_{[3.1]}$  & $\bm{15.1}_{[0.8]}$ & $45\%$  & $150$ \\

Dragon &  $0.1_{[0.06]}$ & $0.1_{[0.1]}$ & $0.3_{[0.3]}$ & $0.3_{[0.3]}$ & $0.1_{[0.12]}$  & $\bm{0.3}_{[0.2]}$ & $200\%$  & $25$ \\

Ztuu &  $3.8_{[0.18]}$ & $4.4_{[0.0]}$ & $4.5_{[0.2]}$ & $4.4_{[0.1]}$ & $4.3_{[0.1]}$  & $\bm{4.5}_{[0.1]}$ & $18\%$  & $100$ \\

\midrule
Norm Score &		$20.1$\% & $19.1$\% &	 $20.6$\% & $20.9$\% & $20.7$ \%  &	$\bm{24.0}$\% & $52.37$\% & $100$\% \\
%\citeyearpar{Gusfield:97} & \small\verb|\citeyearpar| & \small\verb|\shortcite| \\
\midrule
\end{tabular}
\caption{
From the results, it can be consistently seen that LM-in-the-Loop provides a performance improvement over CALM. Especially, categorizing the transitions with state features (OC) scored the highest with $ \sim 53\%$ improvement over the scores obtained by the baseline model.
}
\label{continual_finetuning}
\end{table*}
\subsection{Effect on Performance}
\label{sec:results-performance}
To understand the effect on performance with LM-in-the-Loop, we follow the experimental set up in \S \ref{sec:model-training-details} to evaluate on Jericho benchmark. 
\autoref{table:lm_training} compares the different methods detailed in \S\ref{sec:techniques} with reproduced norm score of CALM \citep{Keep-CALM-Explore-Language-Models-for-Action-Generation-in-Text-based-Games} as the baseline. 
We see that categorizing the transitions using state features ({\bf OC}) scored the highest in all tasks, suggesting that LM-in-the-Loop enables improved performance. This was also reflected in the avg. norm score with an improvement of $\approx 4\%$ over the baseline. This is $\approx 53\%$ more avg. improvement over the scores obtained by the baseline model. Although the performances of {\bf $\mathrm{OC}$} are closer to the baseline in many games, the in-game training accelerated the convergence in most games.

However, the improvement with $\mathrm{\bm{OC}}$ is, in a way, a loose upperbound to in-game learning with LM-in-the-Loop, as special techniques to reweight the transitions ({\bf $\mathrm{UT^{LA}}$}, and {\bf $\mathrm{UT}^{EA}$}), or reward based categorization {\bf $\mathrm{RT}$} only improved the avg. norm score by $\approx 0.6\%$.  On the other hand, the avg. norm score with Uncategorized Transitions ({\bf $\mathrm{UT}$}) dropped to $19.2\%$ which is $\sim 1\%$ below the baseline performance. The difference in performance between ${\bf \mathrm{UT}}$, and ${\bf \mathrm{OC}}$ with the baseline suggests that LM-in-the-loop for action recommendation is helpful but requires careful selection of transitions for training the language model. 
 \begin{figure}[t] 
\includegraphics[width=\columnwidth]{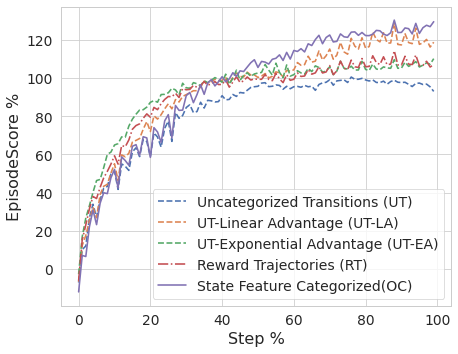}
\caption{We see that LM-in-the-Loop techniques only need half of the steps to achieve the best of CALM. Whereas, using state feature based categorization (OC)  achieved better acceleration and performance over the rest.}
%we see that using reward signals for finetuning provides an accelerated convergence over the baseline.}
\label{fig:acceleration-all}
\end{figure}

 In \autoref{fig:acceleration-all}, we compare the \% of steps in-game learning methods took in average to achieve $k\%$ of CALM model's best performance across the games. We see that LM-in-the-Loop techniques enabled atleast $2\times$ on average\footnote{Individual comparison of each method across the games is in \S\ref{sec:acceleration-plots}.} acceleration in convergence, although the weaker alternatives to $\mathrm{\bm{OC}}$ with reward based categorization, and reweighted techniques only provided meagre improvements over the baseline (\autoref{table:lm_training}).  This shows that the adaptation offered with the ClubFloyd dataset was insufficient, and off-the-shelf techniques can drastically accelerate convergence.

\subsection{Emphasis on Human Annotations}

CALM model---the baseline--- uses all of the $\sim 220K$ transitions in the ClubFloyd dataset to adapt GPT-2 model for action recommendation. But, by using in-game transitions for LM-in-the-Loop training, the LM is provided with game specific information. So, the requirement for adapting GPT-2 with human annotated transitions should be minimal. \citet{Keep-CALM-Explore-Language-Models-for-Action-Generation-in-Text-based-Games} show that CALM's performance decreased significantly when adaptation was done with $10\%$ of ClubFloyd dataset. The reproduced results of CALM with $10\%$ of adaptation data shows the avg. norm score as $18.5\%$  across the games in \autoref{tab:data-efficiency}. Using State features  ({\bf OC}) with $10\%$ of the adaptation date achieved an average norm score of $21.8\%$, which was more than even using $100\%$ of the adaptation data with CALM. Although there was a small decline in the performance of the detective game, it was insignificant because it was still within the standard error. These results suggest empirically that we can reduce the burden of collecting human-played or human-annotated data by doing in-game learning. 

\begin{table}[ht]
\centering
\small
\begin{tabular}{ll|ll}
\toprule
\textbf{Games} & \textbf{CALM }& \textbf{CALM } & \textbf{OC} \\

 & \textbf{100\%}  &  \textbf{10\%}  & \textbf{10\%}\\

\midrule
Zork1       &  $30.7_{[4.8]}$  & $29_{[3.4]}$ &  $\bm{35.1}_{[2.3] }$   \\

Inhumane    &	$24.8_{[2.7]}$   &	 $15.7_{[14.7] } $            & $\bm{27.5}_{[ 6.8] }$   \\
Detective	& 	$290.9_{[2.7]}$  &  $289.5_{[0.2] }$            & $\bm{289.6}_{[0.2 ]}$  \\
Zork3       &   $0.3_{[0.09]}$   &	 $0.6_{[0 ]}$               & $\bm{0.7}_{[0.3]} $   \\
Omniquest   &   $6.7_{[0.3]}$   &	  $5.9_{[0.8 ]}$          & $\textbf{6.0}_{[1 ]}$    \\
Library	    &   $11.2_{[1.3]}$  &  $\bm{10.5}_{[1.5] }$       & $10.2_{[1.8 ]}$  \\
Balances	&  $9.3_{[0.2]} $   &  $6.6_{[ 3.5] }$              & $\bm{8.6}_{[1.6] } $  \\
Ludicorp	&  $10.4_{[0.7]}$  &  $10.2_{[0.4 ]} $              & $\bm{13.7}_{[0.4] } $\\
Dragon	    &  $0.1_{[0.06]}$ &  $0.1 _{[0.06]}$             & $\bm{0.3}_{[0.2 ]} $  \\
Ztuu	    &  $3.8_{[0.18]}$  & $3.6_{[0.1] } $          & $\bm{4.1}_{[0.1] }$    \\
\midrule
Norm &		$20.1$\% &		$18.5$\%	& $\bm{21.8}$ \% \\
\bottomrule
\end{tabular}
\caption{Using State Features (OC) achieved an average norm score
of $21.8\%$ with $10\%$, which was more than even with CALM using $100\%$ of the adaptation data.}
\label{tab:data-efficiency}
\end{table}

\subsection{Effect of Weight Adjusted LM Loss}

Categorization of transitions, although possible in most games, often requires game specific functions to identify what is a good and a bad transition. However, a generalized technique would be to use a notion of the usefulness of transitions that don't require game specific mechanisms. We explore reweighted cross entropy loss as in \autoref{eqn:cross_entropy} with variations of the $h(\cdot)$ functions from being uniformly distributed as $1$ over $(o, a) \in \mathcal{O}\times\mathcal{A}$ to using advantage function with two variations as in \autoref{eqn:weighted-ea} and \autoref{eqn:weighted-la}. While $\mathrm{\bm{UT}}$ uses vanilla cross-entropy loss to train the LM on transitions sampled from buffer $\mathcal{D}$, $\mathrm{\bm{UT^{EA}}}$ and $\mathrm{\bm{UT^{LA}}}$  adjusts the experience according to the advantage, $A(o,a)$, of the actions chosen in those observations.

We use causal language modeling to train the GPT-2 LM to discourage the LM in generating a useful action in a state and discouraging the not useful. 
%n.
As $A(o,a) \in \left[-\infty,+\infty\right]$, it is important to understand how it affects the language model. A negative advantage for $a'$ in $o'$ should discourage the LM from suggesting $a'$ in $o'$. $\mathrm{\bm{UT^{EA}}}$ re-scales the LM-loss with $h(\cdot)\in [0,1)$,  while $\mathrm{\bm{UT^{LA}}}$ works similar to \emph{Unlikelihood} training as proposed in \citet{welleck2019neural} by maintaining the same scale as $A(o,a)$. But, from the restuls we see that the differences in reweighting did not tangible  affect the performance as seen in \autoref{table:lm_training} (Columns $\mathrm{\bm{UT^{EA}}}$ and $\mathrm{\bm{UT^{LA}}}$). %Also, the two methods performed slightly better than the avg. norm score of baseline and provided significant acceleration in convergence, as seen in \autoref{fig:acceleration-all}. 

\subsection{GPT-2 as Policy Network?}

So far, we have explored the performance of LM-in-the-Loop training of GPT-2 for suggesting candidate actions for the DRRN, but to disambiguate the role of GPT-2 and DRRN, we conduct an ablation experiment. Instead of providing action candidates to DRRN agent, what if GPT-2 chose the \emph{argmax} action? The experiment addresses two questions: (1) Is the improvement in the performance and acceleration as in \S \ref{sec:results-performance} largely from the GPT-2 training? and (2) Does the max action of the LM reflect the game dynamics?

\begin{table}[ht]
\centering
\small
\begin{tabular}{l|l|l}
\toprule

\textbf{Games} & \textbf{Frozen LM }& \textbf{In-game LM } \\

\midrule
Zork1       &  $3.3_{[5.7]}$  & $3.3_{[5.7]}$   \\

Inhumane    &	$0_{[0]}$   &	 $0_{[0] } $        \\
Detective	& 	$23.3_{[5.7]}$  &  $15_{[7] }$    \\
Zork3       &   $0_{[0]}$   &	 $0_{[0]}$      \\
Omniquest   &   $0_{[0]}$   &	  $1.6_{[2.8 ]}$        \\

Library	    &   $0_{[0]}$  &  $0_{[0] }$     \\
Balances	&  $0_{[0]} $   &  $0_{[0] }$             \\
Ludicorp	&  $5.3_{[2.0]}$  &  $4.1_{[2.9 ]} $         \\
Dragon	    &  $0_{[0.0]}$ &  $0_{[0.]}$              \\
Ztuu	    &  $0_{[0.0]}$  & $0_{[0.0] } $           \\
\midrule
Norm &		$1.1$\% &		$1.1$\% \\
\bottomrule
\end{tabular}
\caption{
Irrespective of whether the LM was maintained frozen or trained with LM-in-the-Loop, GPT-2 model as policy network yielded zero in the majority of games when DRRN is not used for decision-making.
}
\label{tab:gpt-2-policy-network}
\end{table}

\citet{mcclelland2020placing} motivate the set up of a language model placed in situated learning set up, where it can interact and learn from the environment. However, other than the study conducted in this work, there exists little evidence for interactive learning of an LM from the game transition. \autoref{tab:gpt-2-policy-network} shows the results of how domain adapted pretrained GPT-2 fairs in the ultimate goal of learning solely from interaction on the text games. We observe that the model's performance is $0$ in most games when not using DRRN for decision making. Irrespective of whether the LM was kept frozen or trained with in-game transitions, there was no palpable evidence of language understanding through game semantics observed. 

But, the possible explanation for the performance in \S \ref{sec:results-performance} is that the language model learns more game specific actions, though not optimal, leading to DRRN contributing significantly to the performance observed.

\subsection{Generalization to Other Games}

We observed from the previous results that the agent performing well could be attributed to the actions suggested by the LM that that adapted from the transitions in-game. While that is encouraging, it also risks the generality of such an agent in being transferrable to other games. To quantify the loss in generality, we use the LM-in-the-Loop trained GPT-2 from \emph{zork1} game and continue to train with $4$ different target games---\emph{zork3}, \emph{Ludicorp}, \emph{inhumane}, and \emph{Ztuu}. 

\begin{table*}[ht]
\centering
\small
\begin{tabular}{l|l|ll||l|ll}
\toprule

 & \textbf{ Target}& \textbf{$\mathcal{A}_{\approx}$} & \textbf{$\left(\mathcal{A}\times\mathcal{O}\right)$$_{\approx}$}   & \textbf{CALM} & \textbf{UT} & \textbf{OC} \\

\midrule
\multirow{4}{*}{\rotatebox[origin=c]{90}{zork1}}&         Zork3 &  $0.20$  & $0.056$ & $0.3_{[0.0]}$    & $0_{[0]}$    & $0.3_{[0.3]}$  \\
& ludicorp   &  $ 0.03$ & $0.015$    & $10.4_{[0.7]}$ &   $6.0_{[1.3]}$  &  $11.9_{[2.1]}$   \\
         & inhumane & $0.03$ & $0.026$ & $24.8_{[2.7]}$ & $0_{[0]}$   & $0_{[0]}$  \\
         & Ztuu & $0.04$ & $0.015$  & $3.8_{[0.1]}$  & $0_{[0]}$  &  $1_{[2.1]}$    \\

\bottomrule
% Norm Score &		10.4\% &		9.8\% & 14.8\% & 14.2\% \\
\hline
\end{tabular}
\caption{
Transferring LM-in-the-Loop trained GPT-2 did not provide guarantee improvements over the baseline. Also, the performance remained unexplainable with $\mathcal{A}_{\approx}$ and $(\mathcal{A}\times\mathcal{O})_{\approx}$.}
\label{tab:generality-experiment}
\end{table*}

For the settings to compare, we used using state features ($\mathrm{\bm{OC}}$), and Uncategorized Transitions ($\mathrm{\bm{UT}}$). To train the LM model, we set the buffer size as $100K$ in both the settings. %For $\mathrm{\bm{OC}}$ we set $p^{+}$ based on the values form [0.2,0.4,0.6,0.8,1].
The results tabled in \autoref{tab:generality-experiment} shows that in-game learning techniques suffered from extending to other games when compared with the baseline performance of CALM. As LM-in-the-Loop training performed well on these games when trained in isolation as seen from \autoref{table:lm_training}, it would be interesting to observe if that depended on some notion of similarity between the source and the target games. 

To that end, we define two measures of similarity using the actions, $\mathcal{A}_{\approx}$, and the action-observation combinations $(\mathcal{O}\times\mathcal{A})_{\approx}$ with the target games. For $\mathcal{A}_{\approx}$ we populate the possible actions available from the Jericho environment on the source game, \emph{zork1}, and each of the target games considered. We estimate the BLEU-$2$\cite{papineni-etal-2002-bleu}
score for every action in the source, $a\in\mathcal{A}_s$, with all actions in the target game, $\mathcal{A}_t$, as the reference. The average over the corpus BLEU is tabled in the $\mathcal{A}_{\approx}$ column in \autoref{tab:generality-experiment}. But, $\mathcal{A}_{\approx}$ did not have any reasonable correlation over the performance observed. While \emph{zork3} had an expected action space similarity with the source game and did not have a significant performance drop, the counterfactual scenarios for when $\mathcal{A}_\approx$ is much lower in other games, the performance was mixed. Although in inhumane and Ztuu, $\mathrm{\bm{OC}}$ performed significantly lower than CALM, in Ludicorp the performance was better than the baseline rendering the action similarity score, $\mathcal{A}_{\approx}$, ineffective in explaining the results.
Also, the \textsc{Mauve} score measured between the game transitions, $(\mathcal{O}\times\mathcal{A})_{\approx}$, was too low suggesting a weak semantic overlap between the game spaces. If generality were to be affected with lower similarity between the source and target games measured along action, action$\times$observations, the results were inconsistent in that regard. 

Although \citet{yao2021reading} observed that the models did not naturally respect the notion of semantics, the results that the learnability in the target games being strongly affected when the LM-in-the-Loop is adapted to \emph{zork1} does not entail the LMs being agnostic to notions of semantics. At the same time, the results we observed doesn't either suggest that semantics guide the results. Such mixed observations probably only suggest that notions of semantics through automatic evaluations are not the tools to interpret LMs in text games. %We excluded reward based categorization as the model was not generating rewarding trajectories f
%The average score of the final 100 episodes during training achieved in each game for various models with standard error is shown in \ref{continual_finetuning}, along with the maximum score. Additionally, the average normalized score, the raw score divided by the overall game score, is reported.

\section{Discussion}

The comparison of LM-in-the-Loop with baseline and their absolute performances from \autoref{continual_finetuning} shows that there is more room for improvement. Despite the LMs having strong linguistic priors from pretraining, the large action space when it comes to generative task is one of the significant challenges in adapting LMs to text-based games. Although interactive learning is promising, towards realizing interactive task solving agents, it is imperative to address the issues due to scalability and data-efficiency. The results in the paper through exploring the possibility of adapting language models for action suggestions through utilizing the in-game generated transitions opens up discussions on several key questions: 

While there is improvement in performance, and acceleration in comparison to not learning from the game transitions, the absolute improvement with respect to the games has still a long way to go. When DRRN module was plugged out for ablation, the argmax action of LM was not even close to a reasonable performance indicating the heavy lifting in planning was from DRRN. Towards realizing LMs in situated learning environments,  adapting LMs to different games is a challenging language understanding milestone. Specifically, it is important to align LM's action generation likelihood to reflect the action value function.

Despite the acceleration and a reduced need for human transitions to adapt LMs for action suggestion, interpreting their performance through the conventional lens of automatic semantic and syntax scores is less effective. It is, then, only imperative to make the application of LMs in text games interpretable through automatic metrics that identifies important transitions to train LM-in-the-Loop.
%define a notion of game similarity that reliably explains the results.

%This paper proposed using the non-optimal game transition to enable data-efficient training of language models. Our key insight in this work is that using in-game data with important transitions from the environment has useful information to inform the language model. In exchange, the language model can suggest more relevant actions to the agent to learn an optimized policy to solve the game. Also, in-game learning is extremely data-efficient. We empirically show that the proposed methods alleviate the need for human-annotated data. It could surpass the baseline with one-tenth of human-annotated data, making it extremely useful to quickly transfer to other games or possibly real-world scenarios. 

%Using in-game learning with important transitions, we outperformed \texttt{9 out of 10 games}. We find that the proposed methods reduce the need for human-annotated data. Even with only $10\%$ of human data, we outperformed the norm score by $1.7\%$ compared to the baseline of $100\%$ human data across $10$ games.\\

\section*{Limitations}

The paper analyzes the possibility and challenges in LM-in-the-Loop training of GPT-2 model for action recommendation in text based games. The claims in the work can be further supported with experiments on different LLM. Similarly, the generalization experiments could have added more support to the lack of evidence with additional games. However, these are compute intensive experiments and the claims are largely made in consideration to the limitations in the set up.

\section*{Acknowledgements}

Sarath Chandar is supported by a Canada CIFAR
AI Chair and an NSERC Discovery Grant. The authors acknowledge the
computational resources provided by the Digital
Research Alliance of Canada and Mila Compute resources. We are thankful
to Siva Reddy for
their helpful feedback in this work.

%\section*{Acknowledgments}
%We thank Compute Canada and Mila for providing the computational resources used in this work.  We also thank Siva Reddy, Aristides Milios, Nicholas Meade, Xing Han Lu for peer review.

 % Add footnotes for compute canada and Mila
\bibliography{anthology,acl2020}

\begin{thebibliography}{41}
\expandafter\ifx\csname natexlab\endcsname\relax\def\natexlab#1{#1}\fi

\bibitem[{Ahn et~al.(2022)Ahn, Brohan, Brown, Chebotar, Cortes, David, Finn,
  Fu, Gopalakrishnan, Hausman, Herzog, Ho, Hsu, Ibarz, Ichter, Irpan, Jang,
  Ruano, Jeffrey, Jesmonth, Joshi, Julian, Kalashnikov, Kuang, Lee, Levine, Lu,
  Luu, Parada, Pastor, Quiambao, Rao, Rettinghouse, Reyes, Sermanet, Sievers,
  Tan, Toshev, Vanhoucke, Xia, Xiao, Xu, Xu, Yan, and
  Zeng}]{grounding_language}
Michael Ahn, Anthony Brohan, Noah Brown, Yevgen Chebotar, Omar Cortes, Byron
  David, Chelsea Finn, Chuyuan Fu, Keerthana Gopalakrishnan, Karol Hausman,
  Alex Herzog, Daniel Ho, Jasmine Hsu, Julian Ibarz, Brian Ichter, Alex Irpan,
  Eric Jang, Rosario~Jauregui Ruano, Kyle Jeffrey, Sally Jesmonth, Nikhil~J
  Joshi, Ryan Julian, Dmitry Kalashnikov, Yuheng Kuang, Kuang-Huei Lee, Sergey
  Levine, Yao Lu, Linda Luu, Carolina Parada, Peter Pastor, Jornell Quiambao,
  Kanishka Rao, Jarek Rettinghouse, Diego Reyes, Pierre Sermanet, Nicolas
  Sievers, Clayton Tan, Alexander Toshev, Vincent Vanhoucke, Fei Xia, Ted Xiao,
  Peng Xu, Sichun Xu, Mengyuan Yan, and Andy Zeng. 2022.
\newblock \href {https://doi.org/10.48550/ARXIV.2204.01691} {Do as i can, not
  as i say: Grounding language in robotic affordances}.

\bibitem[{Ammanabrolu and Hausknecht(2020)}]{kga2c}
Prithviraj Ammanabrolu and Matthew Hausknecht. 2020.
\newblock \href {https://openreview.net/forum?id=B1x6w0EtwH} {Graph constrained
  reinforcement learning for natural language action spaces}.
\newblock In \emph{International Conference on Learning Representations}.

\bibitem[{Bender and Koller(2020)}]{climbing_right_hill}
Emily~M. Bender and Alexander Koller. 2020.
\newblock \href {https://doi.org/10.18653/v1/2020.acl-main.463} {Climbing
  towards {NLU}: {On} meaning, form, and understanding in the age of data}.
\newblock In \emph{Proceedings of the 58th Annual Meeting of the Association
  for Computational Linguistics}, pages 5185--5198, Online. Association for
  Computational Linguistics.

\bibitem[{Biewald(2020)}]{wandb}
Lukas Biewald. 2020.
\newblock \href {https://www.wandb.com/} {Experiment tracking with weights and
  biases}.
\newblock Software available from wandb.com.

\bibitem[{Bisk et~al.(2020)Bisk, Holtzman, Thomason, Andreas, Bengio, Chai,
  Lapata, Lazaridou, May, Nisnevich, Pinto, and
  Turian}]{experience_grounds_language}
Yonatan Bisk, Ari Holtzman, Jesse Thomason, Jacob Andreas, Yoshua Bengio, Joyce
  Chai, Mirella Lapata, Angeliki Lazaridou, Jonathan May, Aleksandr Nisnevich,
  Nicolas Pinto, and Joseph Turian. 2020.
\newblock \href {https://doi.org/10.18653/v1/2020.emnlp-main.703} {Experience
  grounds language}.
\newblock In \emph{Proceedings of the 2020 Conference on Empirical Methods in
  Natural Language Processing (EMNLP)}, pages 8718--8735, Online. Association
  for Computational Linguistics.

\bibitem[{Brown et~al.(2020)Brown, Mann, Ryder, Subbiah, Kaplan, Dhariwal,
  Neelakantan, Shyam, Sastry, Askell, Agarwal, Herbert-Voss, Krueger, Henighan,
  Child, Ramesh, Ziegler, Wu, Winter, Hesse, Chen, Sigler, Litwin, Gray, Chess,
  Clark, Berner, McCandlish, Radford, Sutskever, and
  Amodei}]{few_short_learners}
Tom Brown, Benjamin Mann, Nick Ryder, Melanie Subbiah, Jared~D Kaplan, Prafulla
  Dhariwal, Arvind Neelakantan, Pranav Shyam, Girish Sastry, Amanda Askell,
  Sandhini Agarwal, Ariel Herbert-Voss, Gretchen Krueger, Tom Henighan, Rewon
  Child, Aditya Ramesh, Daniel Ziegler, Jeffrey Wu, Clemens Winter, Chris
  Hesse, Mark Chen, Eric Sigler, Mateusz Litwin, Scott Gray, Benjamin Chess,
  Jack Clark, Christopher Berner, Sam McCandlish, Alec Radford, Ilya Sutskever,
  and Dario Amodei. 2020.
\newblock \href
  {https://proceedings.neurips.cc/paper/2020/file/1457c0d6bfcb4967418bfb8ac142f64a-Paper.pdf}
  {Language models are few-shot learners}.
\newblock In \emph{Advances in Neural Information Processing Systems},
  volume~33, pages 1877--1901. Curran Associates, Inc.

\bibitem[{Chen et~al.(2021)Chen, Lu, Rajeswaran, Lee, Grover, Laskin, Abbeel,
  Srinivas, and Mordatch}]{decision_transformer}
Lili Chen, Kevin Lu, Aravind Rajeswaran, Kimin Lee, Aditya Grover, Michael
  Laskin, Pieter Abbeel, Aravind Srinivas, and Igor Mordatch. 2021.
\newblock \href {https://openreview.net/forum?id=a7APmM4B9d} {Decision
  transformer: Reinforcement learning via sequence modeling}.
\newblock In \emph{Advances in Neural Information Processing Systems}.

\bibitem[{Chung et~al.(2014)Chung, Gulcehre, Cho, and Bengio}]{gru}
Junyoung Chung, Caglar Gulcehre, Kyunghyun Cho, and Yoshua Bengio. 2014.
\newblock Empirical evaluation of gated recurrent neural networks on sequence
  modeling.
\newblock In \emph{NIPS 2014 Workshop on Deep Learning, December 2014}.

\bibitem[{C{\^{o}}t{\'{e}} et~al.(2018)C{\^{o}}t{\'{e}}, K{\'{a}}d{\'{a}}r,
  Yuan, Kybartas, Barnes, Fine, Moore, Hausknecht, Asri, Adada, Tay, and
  Trischler}]{text-world}
Marc{-}Alexandre C{\^{o}}t{\'{e}}, {\'{A}}kos K{\'{a}}d{\'{a}}r, Xingdi Yuan,
  Ben Kybartas, Tavian Barnes, Emery Fine, James Moore, Matthew~J. Hausknecht,
  Layla~El Asri, Mahmoud Adada, Wendy Tay, and Adam Trischler. 2018.
\newblock \href {http://arxiv.org/abs/1806.11532} {Textworld: {A} learning
  environment for text-based games}.

\bibitem[{Devlin et~al.(2019{\natexlab{a}})Devlin, Chang, Lee, and
  Toutanova}]{bert}
Jacob Devlin, Ming{-}Wei Chang, Kenton Lee, and Kristina Toutanova.
  2019{\natexlab{a}}.
\newblock \href {https://doi.org/10.18653/v1/n19-1423} {{BERT:} pre-training of
  deep bidirectional transformers for language understanding}.
\newblock In \emph{Proceedings of the 2019 Conference of the North American
  Chapter of the Association for Computational Linguistics: Human Language
  Technologies, {NAACL-HLT} 2019, Minneapolis, MN, USA, June 2-7, 2019, Volume
  1 (Long and Short Papers)}, pages 4171--4186. Association for Computational
  Linguistics.

\bibitem[{Devlin et~al.(2019{\natexlab{b}})Devlin, Chang, Lee, and
  Toutanova}]{bert_pretraining}
Jacob Devlin, Ming{-}Wei Chang, Kenton Lee, and Kristina Toutanova.
  2019{\natexlab{b}}.
\newblock \href {https://doi.org/10.18653/v1/n19-1423} {{BERT:} pre-training of
  deep bidirectional transformers for language understanding}.
\newblock In \emph{Proceedings of the 2019 Conference of the North American
  Chapter of the Association for Computational Linguistics: Human Language
  Technologies, {NAACL-HLT} 2019, Minneapolis, MN, USA, June 2-7, 2019, Volume
  1 (Long and Short Papers)}, pages 4171--4186. Association for Computational
  Linguistics.

\bibitem[{Feng et~al.(2021)Feng, Gangal, Wei, Chandar, Vosoughi, Mitamura, and
  Hovy}]{feng2021survey}
Steven~Y Feng, Varun Gangal, Jason Wei, Sarath Chandar, Soroush Vosoughi,
  Teruko Mitamura, and Eduard Hovy. 2021.
\newblock A survey of data augmentation approaches for nlp.
\newblock \emph{arXiv preprint arXiv:2105.03075}.

\bibitem[{Hausknecht et~al.(2020)Hausknecht, Ammanabrolu, Côté, and
  Yuan}]{interactive_fiction_games}
Matthew Hausknecht, Prithviraj Ammanabrolu, Marc-Alexandre Côté, and Xingdi
  Yuan. 2020.
\newblock \href {https://doi.org/10.1609/aaai.v34i05.6297} {Interactive fiction
  games: A colossal adventure}.
\newblock \emph{Proceedings of the AAAI Conference on Artificial Intelligence},
  34(05):7903--7910.

\bibitem[{He et~al.(2016)He, Chen, He, Gao, Li, Deng, and Ostendorf}]{drrn}
Ji~He, Jianshu Chen, Xiaodong He, Jianfeng Gao, Lihong Li, Li~Deng, and Mari
  Ostendorf. 2016.
\newblock \href {https://doi.org/10.18653/v1/P16-1153} {Deep reinforcement
  learning with a natural language action space}.
\newblock In \emph{Proceedings of the 54th Annual Meeting of the Association
  for Computational Linguistics (Volume 1: Long Papers)}, pages 1621--1630,
  Berlin, Germany. Association for Computational Linguistics.

\bibitem[{Janner et~al.(2021)Janner, Li, and Levine}]{rl_one_big_problem}
Michael Janner, Qiyang Li, and Sergey Levine. 2021.
\newblock \href
  {https://proceedings.neurips.cc/paper/2021/file/099fe6b0b444c23836c4a5d07346082b-Paper.pdf}
  {Offline reinforcement learning as one big sequence modeling problem}.
\newblock In \emph{Advances in Neural Information Processing Systems},
  volume~34, pages 1273--1286. Curran Associates, Inc.

\bibitem[{Joulin et~al.(2017)Joulin, Grave, Bojanowski, and
  Mikolov}]{fast_text}
Armand Joulin, Edouard Grave, Piotr Bojanowski, and Tomas Mikolov. 2017.
\newblock \href {https://aclanthology.org/E17-2068} {Bag of tricks for
  efficient text classification}.
\newblock In \emph{Proceedings of the 15th Conference of the {E}uropean Chapter
  of the Association for Computational Linguistics: Volume 2, Short Papers},
  pages 427--431, Valencia, Spain. Association for Computational Linguistics.

\bibitem[{Khashabi et~al.(2020)Khashabi, Min, Khot, Sabharwal, Tafjord, Clark,
  and Hajishirzi}]{qa}
Daniel Khashabi, Sewon Min, Tushar Khot, Ashish Sabharwal, Oyvind Tafjord,
  Peter Clark, and Hannaneh Hajishirzi. 2020.
\newblock \href {https://doi.org/10.18653/v1/2020.findings-emnlp.171}
  {{UNIFIEDQA}: Crossing format boundaries with a single {QA} system}.
\newblock In \emph{Findings of the Association for Computational Linguistics:
  EMNLP 2020}, pages 1896--1907, Online. Association for Computational
  Linguistics.

\bibitem[{Lake and Murphy(2021)}]{word_meaning}
Brenden~M. Lake and Gregory~L. Murphy. 2021.
\newblock Word meaning in minds and machines.
\newblock \emph{Psychological review}.

\bibitem[{Li et~al.(2022)Li, Puig, Paxton, Du, Wang, Fan, Chen, Huang,
  Aky{\"{u}}rek, Anandkumar, Andreas, Mordatch, Torralba, and
  Zhu}]{lm_for_interactive_decision_making}
Shuang Li, Xavier Puig, Chris Paxton, Yilun Du, Clinton Wang, Linxi Fan, Tao
  Chen, De{-}An Huang, Ekin Aky{\"{u}}rek, Anima Anandkumar, Jacob Andreas,
  Igor Mordatch, Antonio Torralba, and Yuke Zhu. 2022.
\newblock Pre-trained language models for interactive decision-making.
\newblock \emph{arXiv}.

\bibitem[{Maas et~al.(2011)Maas, Daly, Pham, Huang, Ng, and
  Potts}]{classification}
Andrew~L. Maas, Raymond~E. Daly, Peter~T. Pham, Dan Huang, Andrew~Y. Ng, and
  Christopher Potts. 2011.
\newblock \href {http://www.aclweb.org/anthology/P11-1015} {Learning word
  vectors for sentiment analysis}.
\newblock In \emph{Proceedings of the 49th Annual Meeting of the Association
  for Computational Linguistics: Human Language Technologies}, pages 142--150,
  Portland, Oregon, USA. Association for Computational Linguistics.

\bibitem[{McClelland et~al.(2020)McClelland, Hill, Rudolph, Baldridge, and
  Sch{\"u}tze}]{mcclelland2020placing}
James~L McClelland, Felix Hill, Maja Rudolph, Jason Baldridge, and Hinrich
  Sch{\"u}tze. 2020.
\newblock Placing language in an integrated understanding system: Next steps
  toward human-level performance in neural language models.
\newblock \emph{Proceedings of the National Academy of Sciences},
  117(42):25966--25974.

\bibitem[{Ouyang et~al.(2022)Ouyang, Wu, Jiang, Almeida, Wainwright, Mishkin,
  Zhang, Agarwal, Slama, Gray, Schulman, Hilton, Kelton, Miller, Simens,
  Askell, Welinder, Christiano, Leike, and Lowe}]{instructgpt}
Long Ouyang, Jeffrey Wu, Xu~Jiang, Diogo Almeida, Carroll Wainwright, Pamela
  Mishkin, Chong Zhang, Sandhini Agarwal, Katarina Slama, Alex Gray, John
  Schulman, Jacob Hilton, Fraser Kelton, Luke Miller, Maddie Simens, Amanda
  Askell, Peter Welinder, Paul Christiano, Jan Leike, and Ryan Lowe. 2022.
\newblock \href {https://openreview.net/forum?id=TG8KACxEON} {Training language
  models to follow instructions with human feedback}.
\newblock In \emph{Advances in Neural Information Processing Systems}.

\bibitem[{Papineni et~al.(2002)Papineni, Roukos, Ward, and
  Zhu}]{papineni-etal-2002-bleu}
Kishore Papineni, Salim Roukos, Todd Ward, and Wei-Jing Zhu. 2002.
\newblock \href {https://doi.org/10.3115/1073083.1073135} {{B}leu: a method for
  automatic evaluation of machine translation}.
\newblock In \emph{Proceedings of the 40th Annual Meeting of the Association
  for Computational Linguistics}, pages 311--318, Philadelphia, Pennsylvania,
  USA. Association for Computational Linguistics.

\bibitem[{Parisotto et~al.(2019)Parisotto, Song, Rae, Pascanu,
  G{\"{u}}l{\c{c}}ehre, Jayakumar, Jaderberg, Kaufman, Clark, Noury, Botvinick,
  Heess, and Hadsell}]{stable_transformer}
Emilio Parisotto, H.~Francis Song, Jack~W. Rae, Razvan Pascanu, {\c{C}}aglar
  G{\"{u}}l{\c{c}}ehre, Siddhant~M. Jayakumar, Max Jaderberg, Raphael~Lopez
  Kaufman, Aidan Clark, Seb Noury, Matthew~M. Botvinick, Nicolas Heess, and
  Raia Hadsell. 2019.
\newblock \href {http://arxiv.org/abs/1910.06764} {Stabilizing transformers for
  reinforcement learning}.
\newblock \emph{CoRR}, abs/1910.06764.

\bibitem[{Pillutla et~al.(2021)Pillutla, Swayamdipta, Zellers, Thickstun,
  Welleck, Choi, and Harchaoui}]{mauve}
Krishna Pillutla, Swabha Swayamdipta, Rowan Zellers, John Thickstun, Sean
  Welleck, Yejin Choi, and Zaid Harchaoui. 2021.
\newblock Mauve: Measuring the gap between neural text and human text using
  divergence frontiers.
\newblock In \emph{NeurIPS}.

\bibitem[{Radford et~al.(2018{\natexlab{a}})Radford, Narasimhan, Salimans,
  Sutskever et~al.}]{radford2018improving}
Alec Radford, Karthik Narasimhan, Tim Salimans, Ilya Sutskever, et~al.
  2018{\natexlab{a}}.
\newblock Improving language understanding by generative pre-training.

\bibitem[{Radford et~al.(2018{\natexlab{b}})Radford, Wu, Child, Luan, Amodei,
  and Sutskever}]{gpt2}
Alec Radford, Jeffrey Wu, Rewon Child, David Luan, Dario Amodei, and Ilya
  Sutskever. 2018{\natexlab{b}}.
\newblock \href
  {https://d4mucfpksywv.cloudfront.net/better-language-models/language-models.pdf}
  {Language models are unsupervised multitask learners}.

\bibitem[{Rajpurkar et~al.(2016)Rajpurkar, Zhang, Lopyrev, and Liang}]{squad}
Pranav Rajpurkar, Jian Zhang, Konstantin Lopyrev, and Percy Liang. 2016.
\newblock \href {https://doi.org/10.18653/v1/D16-1264} {{SQ}u{AD}: 100,000+
  questions for machine comprehension of text}.
\newblock In \emph{Proceedings of the 2016 Conference on Empirical Methods in
  Natural Language Processing}, pages 2383--2392, Austin, Texas. Association
  for Computational Linguistics.

\bibitem[{Reid et~al.(2022)Reid, Yamada, and Gu}]{offline-wikipedia}
Machel Reid, Yutaro Yamada, and Shixiang~Shane Gu. 2022.
\newblock \href {http://arxiv.org/abs/2201.12122} {Can wikipedia help offline
  reinforcement learning?}
\newblock \emph{CoRR}, abs/2201.12122.

\bibitem[{Rogers et~al.(2020)Rogers, Kovaleva, and
  Rumshisky}]{rogers2020primer}
Anna Rogers, Olga Kovaleva, and Anna Rumshisky. 2020.
\newblock A primer in bertology: What we know about how bert works.
\newblock \emph{Transactions of the Association for Computational Linguistics},
  8:842--866.

\bibitem[{Singh et~al.(2021)Singh, Singh, and Modi}]{bert_textgame}
Ishika Singh, Gargi Singh, and Ashutosh Modi. 2021.
\newblock \href {http://arxiv.org/abs/2107.08408} {Pre-trained language models
  as prior knowledge for playing text-based games}.
\newblock \emph{CoRR}, abs/2107.08408.

\bibitem[{Tarasov et~al.(2022)Tarasov, Kurenkov, and
  Kolesnikov}]{LM_for_offline_rl}
Denis Tarasov, Vladislav Kurenkov, and Sergey Kolesnikov. 2022.
\newblock \href {https://openreview.net/forum?id=Spf4TE6NkWq} {Prompts and
  pre-trained language models for offline reinforcement learning}.
\newblock In \emph{ICLR 2022 Workshop on Generalizable Policy Learning in
  Physical World}.

\bibitem[{Tuyls et~al.(2022)Tuyls, Yao, Kakade, and
  Narasimhan}]{Multi-Stage-Episodic-Control-for-Strategic-Exploration-in-Text-Games}
Jens Tuyls, Shunyu Yao, Sham~M. Kakade, and Karthik~R Narasimhan. 2022.
\newblock \href {https://openreview.net/forum?id=Ek7PSN7Y77z} {Multi-stage
  episodic control for strategic exploration in text games}.
\newblock In \emph{International Conference on Learning Representations}.

\bibitem[{Webson and Pavlick(2021)}]{webson2021prompt}
Albert Webson and Ellie Pavlick. 2021.
\newblock Do prompt-based models really understand the meaning of their
  prompts?
\newblock \emph{arXiv preprint arXiv:2109.01247}.

\bibitem[{Welleck et~al.(2019)Welleck, Kulikov, Roller, Dinan, Cho, and
  Weston}]{welleck2019neural}
Sean Welleck, Ilia Kulikov, Stephen Roller, Emily Dinan, Kyunghyun Cho, and
  Jason Weston. 2019.
\newblock Neural text generation with unlikelihood training.
\newblock \emph{arXiv preprint arXiv:1908.04319}.

\bibitem[{Wolf et~al.(2020)Wolf, Debut, Sanh, Chaumond, Delangue, Moi, Cistac,
  Rault, Louf, Funtowicz, Davison, Shleifer, von Platen, Ma, Jernite, Plu, Xu,
  Le~Scao, Gugger, Drame, Lhoest, and Rush}]{transformer_training}
Thomas Wolf, Lysandre Debut, Victor Sanh, Julien Chaumond, Clement Delangue,
  Anthony Moi, Pierric Cistac, Tim Rault, Remi Louf, Morgan Funtowicz, Joe
  Davison, Sam Shleifer, Patrick von Platen, Clara Ma, Yacine Jernite, Julien
  Plu, Canwen Xu, Teven Le~Scao, Sylvain Gugger, Mariama Drame, Quentin Lhoest,
  and Alexander Rush. 2020.
\newblock \href {https://doi.org/10.18653/v1/2020.emnlp-demos.6} {Transformers:
  State-of-the-art natural language processing}.
\newblock In \emph{Proceedings of the 2020 Conference on Empirical Methods in
  Natural Language Processing: System Demonstrations}, pages 38--45, Online.
  Association for Computational Linguistics.

\bibitem[{Wu et~al.(2022)Wu, Li, and Liang}]{wu2022insights}
Yuhuai Wu, Felix Li, and Percy Liang. 2022.
\newblock Insights into pre-training via simpler synthetic tasks.
\newblock \emph{arXiv preprint arXiv:2206.10139}.

\bibitem[{Xu et~al.(2020)Xu, Chen, Fang, Wang, and
  Zhang}]{DRL_with_transformers}
Yunqiu Xu, Ling Chen, Meng Fang, Yang Wang, and Chengqi Zhang. 2020.
\newblock \href {https://doi.org/10.1109/CoG47356.2020.9231622} {Deep
  reinforcement learning with transformers for text adventure games}.
\newblock In \emph{2020 IEEE Conference on Games (CoG)}, pages 65--72.

\bibitem[{Yao et~al.(2021)Yao, Narasimhan, and Hausknecht}]{yao2021reading}
Shunyu Yao, Karthik Narasimhan, and Matthew Hausknecht. 2021.
\newblock Reading and acting while blindfolded: The need for semantics in text
  game agents.
\newblock In \emph{Proceedings of the 2021 Conference of the North American
  Chapter of the Association for Computational Linguistics: Human Language
  Technologies}, pages 3097--3102.

\bibitem[{Yao et~al.(2020)Yao, Rao, Hausknecht, and
  Narasimhan}]{Keep-CALM-Explore-Language-Models-for-Action-Generation-in-Text-based-Games}
Shunyu Yao, Rohan Rao, Matthew Hausknecht, and Karthik Narasimhan. 2020.
\newblock \href {https://doi.org/10.18653/v1/2020.emnlp-main.704} {Keep {CALM}
  and explore: Language models for action generation in text-based games}.
\newblock In \emph{Proceedings of the 2020 Conference on Empirical Methods in
  Natural Language Processing (EMNLP)}, pages 8736--8754, Online. Association
  for Computational Linguistics.

\bibitem[{Zhang et~al.(2015)Zhang, Zhao, and LeCun}]{text_classification}
Xiang Zhang, Junbo Zhao, and Yann LeCun. 2015.
\newblock \href
  {https://proceedings.neurips.cc/paper/2015/file/250cf8b51c773f3f8dc8b4be867a9a02-Paper.pdf}
  {Character-level convolutional networks for text classification}.
\newblock In \emph{Advances in Neural Information Processing Systems},
  volume~28. Curran Associates, Inc.

\end{thebibliography}
\bibliographystyle{acl_natbib}

\appendix

\section{Appendix}
\subsection{Language Model Setup}  \label{lm setup}

We use a GPT-2 (Base) \citep{gpt2} model with $12$-layers, $768$-hidden units, and $12$- attention heads with $117$M parameters pre-trained on the WebText corpus. This model's implementation and pretrained weights are obtained from \citep[Huggingface]{transformer_training}.
% clubfloyd 

We train for $3$ epochs on the ClubFloyd dataset following \citep{Keep-CALM-Explore-Language-Models-for-Action-Generation-in-Text-based-Games} to minimize the cross-entropy loss, as shown in Table \ref{table:lm_training}. We use AdamW to optimize model's weights to minimize the loss, with the learning rate as $2\times 10^{-6}$ and Adam epsilon as $1\times10^{-9}$. We use a linear schedule with a warmup of $0.1$ for the learning rate. Finally, we clip gradients with a maximum gradient norm of $1$. Following \cite{Keep-CALM-Explore-Language-Models-for-Action-Generation-in-Text-based-Games}'s finetuning process, we exclude using Jericho-related transcripts by setting the flag as $1$. We used random seeds to select the dataset to avoid bias in selecting data for the LM training.

\begin{table}[ht]
\centering
\begin{tabular}{lll}
\hline

\textbf{Model} & \textbf{Metric}& \textbf{Final Score(3 epoch)} \\
\hline
100\%  & Train Loss & 1.49 \\
  & Val Loss  &  2.65  \\

 & Train Acc  & 0.30  \\
  & Val Acc  & 0.14  \\

\hline
\hline
10\%  & Train Loss & 1.42  \\
  & Val Loss  & 3.04  \\

 & Train Acc  &  0.30 \\

& Val Acc & 0.09  \\
\hline

\end{tabular}
\caption{\label{table:lm_training}
Pre-trained GPT-2 Language Model training details on different data percentage variants trained.
}
\end{table}
%beam search to select 30 action candidates.

% talk about the replay size
% frequency of finetuning
% gradient updates

\subsection{Reinforcement Learning Agent Setup:} 
% number 10 should always be consistent
% Add an image to help you avoid redundant explanations.
%max_step it set to 100.
% use variable that is self-explanatory 
% use texttt for hightlight 
% learning rate and notation consistent
%

We train on $10$ interactive fiction games from the Jericho benchmark \citep{interactive_fiction_games}. The states are observations concatenated with items in possession of the player and their current location description provided by the game engine using commands inventory and look. A single game episode runs for $100$ environment steps at max or gets terminated before the game is over or won.   We use the \texttt{look} and \texttt{inventory} commands to add location and inventory descriptions to observations, following \citet{interactive_fiction_games}. 
% previous practice from 

We train DRRN asynchronously on $8$ parallel instances of the game environment for $100,000$ steps for each game. At each step, the Q-value is estimated using the DRRN agent, and the action is selected based on the soft-exploration policy. Action's admissibility is predicted based on the textual response of the game. Then, inadmissible are filtered out using a FastText model \citep{fast_text}. The agent is optimized using adam optimizer with a learning rate of $10^{-5}$. We sample transitions of batch size $64$ from priority buffer with a priority fraction of $0.5$. The discount factor in determining the future reward's importance is $0.9$. The size of the embedding dimension is $128$, and the hidden dimension is $128$. Finally, the gradient is clipped with a maximum gradient norm of $5$.
We train $5$ separate runs for each game and report the average score. We use the average of the last $100$ episode scores to calculate the final score.

% As the agent interacts with the environment, all the context action pairs ${((o_{t−1}, a_{t−1}, o_{t}), a_{t})}$  are stored in one of the First-In-First-Out (FIFO) replay buffer size $D^{*}$ of $100K$. Important context action pairs, i.e., reward-yielding and location-changing transition, are stored in the second FIFO replay buffer of size $100K$.
\subsection{Software Details}
We used PyTorch for the code implementation and Huggingface to load pre-trained language models. We used Weights \& Biases \citep{wandb} for experiment tracking and visualizations to develop insights for this paper. Finally, the seaborn package is used to generate plots.

\subsection{Mauve Score}
\label{sec:mauve-clubfloyd-jericho}
\begin{table}[ht]
\small
\centering
\begin{tabular}{l|l|l}
\toprule
 & \textbf{ Target}& \textbf{Mauve} \\

\midrule
\multirow{8}{*}{\rotatebox[origin=c]{90}{ClubFloyd}}
        &  Zork1 &  $0.017$     \\
        & Inhumane   &  $ 0.013$       \\
         & Detective & $0.009$   \\
         & Zork3 & $0.014$      \\
         & Omniquest & $0.009$      \\
         & Library & $0.023$      \\
         & Balances & $0.012$      \\
         & Ludicorp & $0.009$      \\
         & Dragon & $0.017$      \\
         & Ztuu & $0.013$      \\

\bottomrule
% Norm Score &		10.4\% &		9.8\% & 14.8\% & 14.2\% \\
\hline
\end{tabular}
\caption{
Mauve score between ClubFloyd and Jericho games}
\label{tab: mauve-jericho-clubfloyd}
\end{table}

\section{Acceleration Plots}
\label{sec:acceleration-plots}

\begin{figure}[h]
\begin{subfigure}[b]{\columnwidth} 
\includegraphics[scale=0.5]{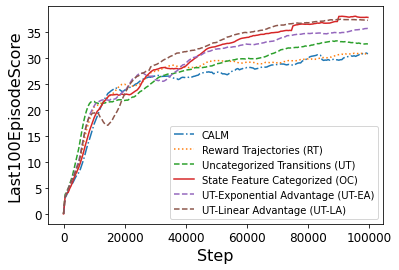}
\subcaption{zork1}
\end{subfigure}

\begin{subfigure}[b]{\columnwidth} 
\includegraphics[scale=0.5]
{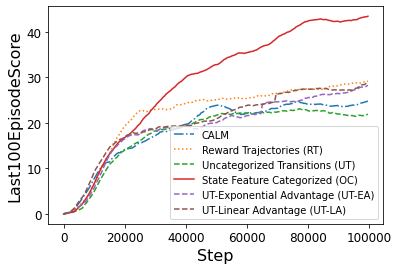}
\subcaption{inhumane}
\end{subfigure}
\end{figure}
\begin{figure}[ht]\ContinuedFloat
\begin{subfigure}[b]{\columnwidth} 
\includegraphics[scale=0.5]
{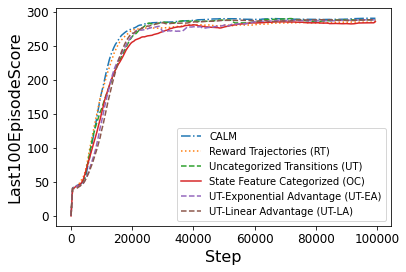}
\caption{detective}
\end{subfigure}

\begin{subfigure}[b]{\columnwidth} 
\includegraphics[scale=0.5]
{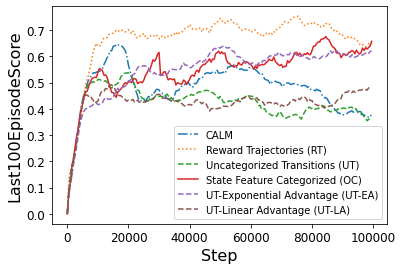}
\caption{zork3}
\end{subfigure}

\begin{subfigure}[b]{\columnwidth} 
\includegraphics[scale=0.5]
{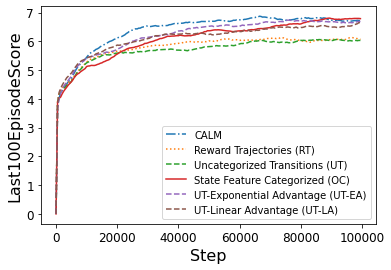}
\caption{omniquest}
\end{subfigure}

\begin{subfigure}[b]{\columnwidth} 
\includegraphics[scale=0.5]{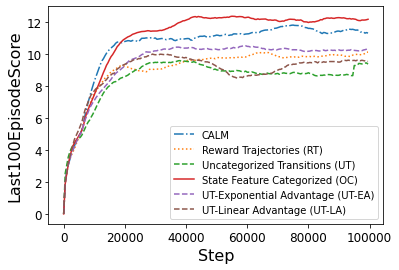}
\caption{library}
\end{subfigure}

\end{figure}
\begin{figure}[ht]\ContinuedFloat
    \begin{subfigure}[b]{\columnwidth} 
\includegraphics[scale=0.5]{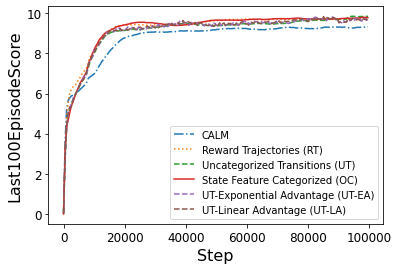}
\caption{balances}
\end{subfigure}

\begin{subfigure}[b]{\columnwidth} 
\includegraphics[scale=0.5]{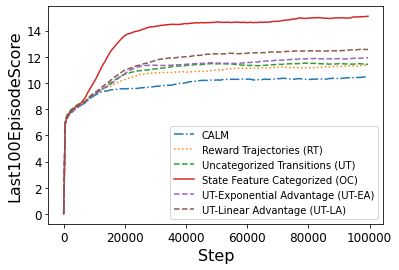}
\caption{ludicorp}
\end{subfigure}

\begin{subfigure}[b]{\columnwidth} 
\includegraphics[scale=0.5]{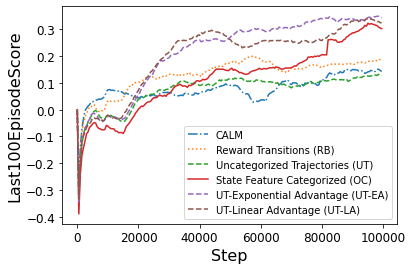}
\caption{dragon}
\end{subfigure}

\begin{subfigure}[b]{\columnwidth} 
\includegraphics[scale=0.5]{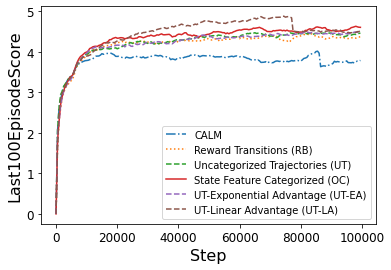}
\caption{ztuu}
\end{subfigure}
\caption{Comparison of learning dynamics of the different LM-in-the-Loop techniques with the baseline CALM agent across the selected $10$ games in Jericho.}
    %\label{fig:my_label}
\end{figure}
\end{document}